\documentclass[letterpaper]{article} 
\usepackage{aaai25}  
\usepackage{times}  
\usepackage{helvet}  
\usepackage{courier}  
\usepackage[hyphens]{url}  
\usepackage{graphicx} 
\usepackage{natbib}  
\usepackage{caption} 
\usepackage{algorithm}
\usepackage{algorithmic}

\usepackage{newfloat}
\usepackage{listings}
\DeclareCaptionStyle{ruled}{labelfont=normalfont,labelsep=colon,strut=off} 
\floatstyle{ruled}
\newfloat{listing}{tb}{lst}{}
\floatname{listing}{Listing}
\usepackage{amsmath}
\usepackage{amsfonts}
\usepackage{booktabs}
\usepackage{cleveref}
\usepackage{makecell}
\usepackage{multirow}
\usepackage{subcaption}
\usepackage{xspace}

\makeatletter
\DeclareRobustCommand\onedot{\futurelet\@let@token\@onedot}
\def\@onedot{\ifx\@let@token.\else.\null\fi\xspace}

\def\eg{\emph{e.g}\onedot} 
\def\ie{\emph{i.e}\onedot}

\makeatother

\newsavebox\CBox
\def\textBF#1{\sbox\CBox{#1}\resizebox{\wd\CBox}{\ht\CBox}{\textbf{#1}}}

\title{Rethinking Transformer-Based Blind-Spot Network\\ for Self-Supervised Image Denoising}
\author{
    Junyi Li, Zhilu Zhang, and Wangmeng Zuo\thanks{Corresponding author.}
}
\affiliations{
    Harbin Institute of Technology, Harbin, China\\

    nagejacob@gmail.com, cszlzhang@outlook.com, wmzuo@hit.edu.cn

}

\usepackage{bibentry}

\begin{document}

\maketitle

\begin{abstract}
Blind-spot networks (BSN) have been prevalent neural architectures in self-supervised image denoising (SSID).
However, most existing BSNs are conducted with convolution layers. 
Although transformers have shown the potential to overcome the limitations of convolutions in many image restoration tasks, the attention mechanisms may violate the blind-spot requirement, thereby restricting their applicability in BSN.
To this end, we propose to analyze and redesign the channel and spatial attentions to meet the blind-spot requirement.
Specifically, channel self-attention may leak the blind-spot information in multi-scale architectures, since the downsampling shuffles the spatial feature into channel dimensions.
To alleviate this problem, we divide the channel into several groups and perform channel attention separately.
For spatial self-attention, we apply an elaborate mask to the attention matrix to restrict and mimic the receptive field of dilated convolution.
Based on the redesigned channel and window attentions, we build a \textbf{T}ransformer-based \textbf{B}lind-\textbf{S}pot \textbf{N}etwork (TBSN), which shows strong local fitting and global perspective abilities.
Furthermore, we introduce a knowledge distillation strategy that distills TBSN into smaller denoisers to improve computational efficiency while maintaining performance. Extensive experiments on real-world image denoising datasets show that TBSN largely extends the receptive field and exhibits favorable performance against state-of-the-art SSID methods.
\end{abstract}

%
\begin{links}
    \link{Code}{https://github.com/nagejacob/TBSN}
\end{links}

\section{Introduction}
Image denoising is a fundamental low-level vision task that aims to recover implicit clean images from their noisy observations.
With the development of convolutional neural networks, learning-based methods~\cite{mao2016image, zhang2017beyond, tai2017memnet, lin2023improving} have shown significant improvements against traditional ones~\cite{buades2005non, dabov2007image}.
In order to facilitate network training, it is common to synthesize noisy-clean image pairs with additive white Gaussian noise (AWGN) for supervised learning.
Since the distribution gap between AWGN and camera noise, they exhibit degraded denoising performance in real-world scenarios.
One feasible solution is to capture datasets~\cite{plotz2017benchmarking, abdelhamed2018sidd} with strictly aligned noisy-clean pairs for network training~\cite{guo2019toward, kim2020transfer}.
However, the data collection process requires rigorously controlled environments and much human labor, which is less practical.

\begin{figure}[t]
	\centering
	\captionsetup{type=figure}
	\begin{subfigure}[h]{0.24\linewidth}
		\includegraphics[width=\linewidth]{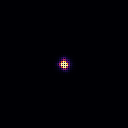}
		\caption*{\centering AP-BSN}
	\end{subfigure}
	\hfill
	\begin{subfigure}[h]{0.24\linewidth}
		\includegraphics[width=\linewidth]{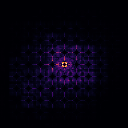}
		\caption*{\centering LG-BPN}
	\end{subfigure}
	\hfill
	\begin{subfigure}[h]{0.24\linewidth}
		\includegraphics[width=\linewidth]{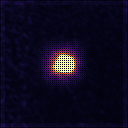}
		\caption*{\centering PUCA}
	\end{subfigure}
	\hfill
	\begin{subfigure}[h]{0.24\linewidth}
		\includegraphics[width=\linewidth]{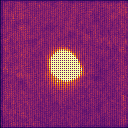}
		\caption*{\centering TBSN (Ours)}
	\end{subfigure}
	\caption{Visualization of effective receptive field for representative blind-spot networks (BSNs). Benefiting from proposed dilated window and channel attention mechanisms, our TBSN shows stronger local fitting and global information aggregation capability with respect to previous BSNs.}
	\label{fig:receptive_field}
\end{figure}

Recently, self-supervised image denoising (SSID)~\cite{krull2019noise2void, lee2022ap, li2023spatially, wang2023lg, jang2023puca} has been introduced to circumvent the requirement of the paired datasets.
The pioneering work Noise2Void randomly masks some locations of a noisy input and trains the network to reconstruct them from their surrounding ones.
In the case of pixel-wise independent noise, the network learns to predict the masked pixels without random noise, \ie, the clean pixels, thereby achieving self-supervised denoising.
Blind-spot networks (BSN)~\cite{laine2019high, wu2020unpaired, byun2021fbi} take a step further to implement the mask schema with dedicated designed network architectures that exclude the corresponding input pixel from the receptive field of each output location, which shows superiority in both performance and training efficiency.
For spatially correlated noise removal in real-world scenarios, some works~\cite{wu2020unpaired, zhou2020awgn} suggest first breaking the noise correlation with pixel-shuffle downsampling (PD), and then denoising with BSN.
In particular, asymmetric PD strategy~\cite{lee2022ap, wang2023lg, jang2023puca} for training and inference has shown a better trade-off between noise removal and detail preserving.

Existing BSN architectures are mostly convolutional neural networks (CNNs).
However, convolution operations have limited capability in capturing long-range dependencies and the static weights of convolution filters could not flexibly adapt to the input content.
As highlighted by the success of the image restoration models~\cite{liang2021swinir, zamir2022restormer, chen2023activating}, such limitations could be mitigated with transformer models~\cite{vaswani2017attention}.
Nonetheless, transformer operators may violate the blind-spot requirement and lead to overfitting to the noisy inputs.
Despite the difficulty, few attempts have been made to apply transformers into BSNs.
For instance, LG-BPN~\cite{wang2023lg} incorporates channel-wise self-attention~\cite{zamir2022restormer} for global feature enhancement but still uses convolution layers for local information integration.
SwinIA~\cite{papkov2023swinia} implements a swin-transformer~\cite{liu2021swin} based BSN with modified window attention. However, limited by the requirement for blind spots, it can only exploit shallow features of the noisy inputs in the attention layer, thus showing inferior performance.
It can be seen that it is very challenging to bring out the effective capabilities of transformers in BSN.

In this paper, we propose to analyze the spatial and channel self-attention mechanisms and redesign them to meet the blind-spot requirement.
For channel-wise self-attention, we observe that simply applying it may leak the blind-spot information, especially in multi-scale architectures.
The deep features of such architectures have been downsampled multiple times and the spatial information is shuffled to the channel dimension.
The interaction between channels may leak spatial information at the blind-spot, leading to overfitting to the noisy input.
We empirically find that this effect appears when the channel dimension is larger than the spatial resolution.
To eliminate the undesirable effect, we divide the channels into groups and perform channel attention on each group separately, where the group channel number is controlled less than spatial resolution.
For spatial self-attention, we elaborately redesign window attention by restricting its receptive field to maintain the blind-spot requirement.
Specifically, a fixed mask is applied to the attention matrix so that each pixel can only attend to pixels at even coordinates.
%
%
Combining the designed spatial and channel self-attention mechanisms, we propose a dilated transformer attention block (DTAB).
We embed DTAB into the encoder-decoder based U-Net architecture, thus presenting a transformer-based blind-spot network (TBSN).

Additionally, BSN architectures are mostly computationally inefficient due to the additional design for satisfying the blind-spot requirement.
It becomes even worse with increasing model size and complicated post-refinement process~\cite{lee2022ap}.
However, some simple and efficient supervised denoisers have the potential to reach the performance of state-of-the-art SSID methods. 
In this work, we take advantage of this property to explore a knowledge distillation strategy for reducing the computation cost during inference.
Specifically, we regard the results of pre-trained TBSN as pseudo ground-truths, and take them as supervision to train a plain U-Net, namely TBSN2UNet.

Extensive experiments are conducted on real-world denoising datasets~\cite{abdelhamed2018sidd, plotz2017benchmarking} to assess the effectiveness of TBSN and TBSN2UNet.
As shown in Fig.~\ref{fig:receptive_field}, benefiting from proposed spatial and channel self-attention mechanisms, TBSN enhances the local adaptivity and largely expands the receptive field.
TBSN behaves favorably against state-of-the-art SSID methods in terms of both quantitative metrics and perceptual quality.
Moreover, TBSN2UNet maintains the performance of TBSN while significantly reducing inference costs.

Our main contributions can be summarized as follows:

\begin{itemize}
	\item We propose a transformer-based blind-spot network (TBSN) that contains spatial and channel self-attentions for self-supervised image denoising.
	\item For channel self-attention, we find it may leak the blind-spot information when the channel number becomes large, we thus perform it on each divided channel group separately to eliminate this adverse effect. For spatial self-attention, we introduce masked window attention where an elaborate mask is applied to the attention matrix to maintain the blind-spot requirement. 
	\item Extensive experiments demonstrate that TBSN achieves state-of-the-art performance on real-world image denoising datasets, while our U-Net distilled from TBSN effectively reduces the computation cost during inference.
\end{itemize}

\section{Related Work}

\subsection{Deep Image Denoising}
The development of learning-based methods~\cite{zhang2017beyond} has shown superior performance against traditional patch-based ones~\cite{buades2005non, dabov2007image} on synthetic Gaussian denoising.
More advanced deep neural architectures~\cite{mao2016image, tai2017memnet, liu2018multi} are further proposed to improve the denoising ability.
NBNet~\cite{cheng2021nbnet} proposes a noise basis network by learning a set of reconstruction basis in the feature space.
InvDN~\cite{liu2021invertible} proposes a lightweight denoising network based on normalizing flow architectures.
Recently, transformers that were first introduced for sequence modeling in natural language processing~\cite{vaswani2017attention} have been successfully applied to vision tasks~\cite{dosovitskiy2020image, liu2021swin}.
For image denoising, transformers are studied with large-scale image pre-training~\cite{chen2021pre} and Swin Transformer architectures~\cite{liang2021swinir}.
Restormer~\cite{zamir2022restormer} and Uformer~\cite{wang2022uformer} propose multi-scale hierarchical network designs, which achieve better trade-offs between performance and efficiency.
However, there are limited efforts for adapting transformers into self-supervised image denoising~\cite{wang2023lg, papkov2023swinia} due to the blind-spot requirement.

\begin{figure*}[t]
    \centering
    \includegraphics[width=0.9\linewidth]{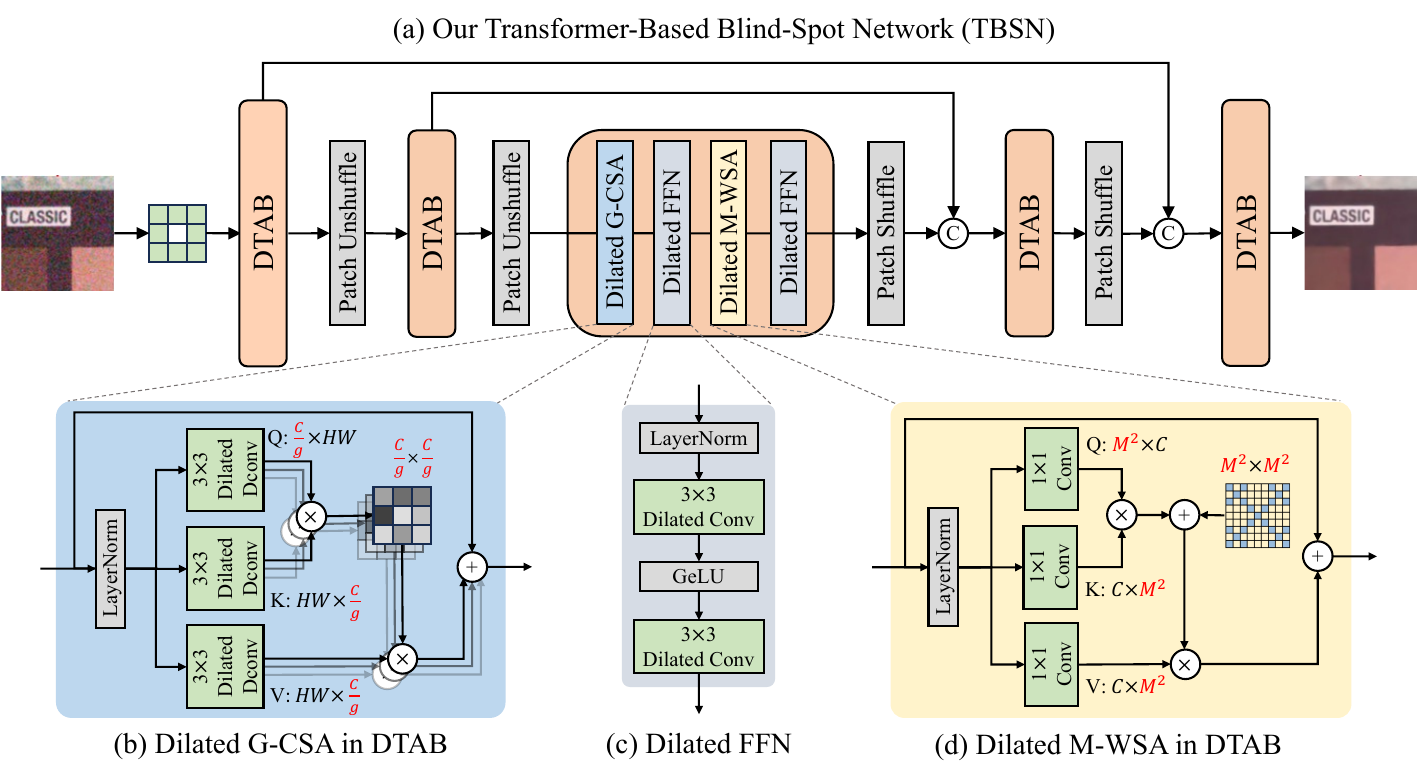}
    \caption{Overview of transformer-based blind-spot network (\textbf{TBSN}). It deploys multi-scale dilated transformer attention blocks (DTAB) to mitigate the shortcomings of convolutions. The core modules of DTAB are dilated counterparts of grouped channel-wise self-attention (G-CSA), masked window-based self-attention (M-WSA), and feed-forward network (FFN), respectively.}
    \label{fig:overview}
\end{figure*}
\subsection{Self-Supervised Image Denoising}
Self-supervised image denoising (SSID) seeks to utilize the information from the noisy images themselves as supervision~\cite{krull2019noise2void, batson2019noise2self}.
In order to prevent trivial solutions such as over-fitting to the identity mapping, blind-spot networks (BSN)~\cite{wu2020unpaired} exclude the corresponding noisy pixel from the receptive field at every spatial location.
Probabilistic inference~\cite{laine2019high} and regular loss functions~\cite{huang2021neighbor2neighbor, zhang2022self} are further introduced to recover the missing information at the blind-spot.
For real-world RGB image denoising, the noise is spatially correlated due to the demosaic operation in image signal processing (ISP) pipeline~\cite{guo2019toward}.
It will easily fit the input noise when deploying BSN designed for spatially independent noise removal.
One feasible solution is to break the noisy correlation with pixel-shuffle downsampling~\cite{zhou2020awgn}, then apply BSN to the downsampled images~\cite{lee2022ap, wang2023lg, pan2023random, jang2023self, jang2023puca}.
In addition, CVF-SID~\cite{neshatavar2022cvf} learns a cyclic function to decompose the noisy image into clean and noisy components.
SASL~\cite{li2023spatially} detects flat and textured areas then constructs supervisions for them separately.
Although much effort has been made for developing SSID algorithms~\cite{cheng2023score, lin2023unsupervised, chen2023multi, zou2023iterative, wang2023noise2info}, there is still a lack of further exploration for BSN architectures.
In this work, we adapt the transformer mechanism to BSN to further unleash the potential of blind-spot manners.

\section{Method}
\subsection{Overview of the Network Architecture}
As shown in Fig.~\ref{fig:overview}, TBSN follows dilated BSN~\cite{wu2020unpaired} to apply 3$\times$3 centrally masked convolution at the first layer and dilated transformer attention blocks (DTABs) in the remaining layers.
The network architecture is U-Net and adopts patch-unshuffle/shuffle~\cite{jang2023puca} based downsampling/upsampling operations to maintain the blind-spot requirement.
The building block, \ie, DTAB, is formed with dilated counterparts of grouped channel-wise self-attention (G-CSA), masked window-based self-attention (M-WSA), and feed-forward network (FFN), respectively.
Thus, TBSN benefits from both the global interaction of channel attention and the local fitting ability of window attention.
We will provide detailed illustrations of the network design in the following subsections.

\subsection{Grouped Channel-Wise Self-Attention (G-CSA)}
\label{sec:ca}
Channel attention~\cite{hu2018squeeze} recalibrates the channel-wise feature responses by explicitly modeling the interdependencies between channels.
Given an input feature $\mathbf{X}\!\in\!\mathbb{R}^{H\times W\times C}$, channel attention can be formalized as,
\begin{equation}
	\text{CA}(\mathbf{X}) = \mathbf{X}\ast  \phi (\mathbf{X}).
	\label{equ:att1}
\end{equation}
where function $\phi (\cdot)$ aggregates the spatial information in each channel, and $\ast$ is channel-wise multiplication operation.
For instance, NAFNet~\cite{chen2022simple} achieves $\phi (\cdot)$ by global average pooling, while Restormer~\cite{zamir2022restormer} applies transposed matrix multiplication in the channel dimension.
However, in SSID task, channel attention may leak blind-spot information as $\phi (\cdot)$ aggregates the content of all the spatial locations, which is ignored in previous methods.

In this work, we systematically analyze the effects of channel attention (CA) in BSN and empirically find it depends on the channel number versus spatial resolution.
For single-level architectures~\cite{wang2023lg}, spatial information is largely compressed by global average pooling, thus CA is beneficial for performance.
For multi-scale architectures~\cite{jang2023puca}, the spatial information is shuffled to various channels by downsampling operations. Thus, CA may be partially equivalent to spatial interaction, leaking the blind-spot values.
To this end, we propose to control the channel number smaller than spatial resolution.
Specifically, we introduce grouped channel-wise self-attention (G-CSA) to divide the deep feature into multiple channel groups and perform CA separately.
Our G-CSA could be formulated as,
\begin{equation}
	\text{G-CSA}(\mathbf{X}) = \text{Concat}(\mathbf{X}_1\ast  \phi (\mathbf{X}_1), \cdots, \mathbf{X}_G\ast  \phi (\mathbf{X}_G)).
	\label{eqn:G-CSA}
\end{equation}
where $\mathbf{X}=\text{Concat}(\mathbf{X}_1, \cdots, \mathbf{X}_G)$, $G$ is the group number.
We set the channel number of each group, \ie, $\frac{C}{G}$, to be small enough to avoid the leakage of spatial information.
In the implementation, we adapt MDTA~\cite{zamir2022restormer} to our G-CSA with Eq.~(\ref{eqn:G-CSA}) for global interaction.
We also replace the $3\times 3$ depth convolutions with their dilated counterparts to achieve blind-spot requirement, as shown in Fig.~\ref{fig:overview}(b).

\begin{figure}[t]
	\centering
	\begin{subfigure}[h]{0.9\linewidth}
		\includegraphics[width=\linewidth]{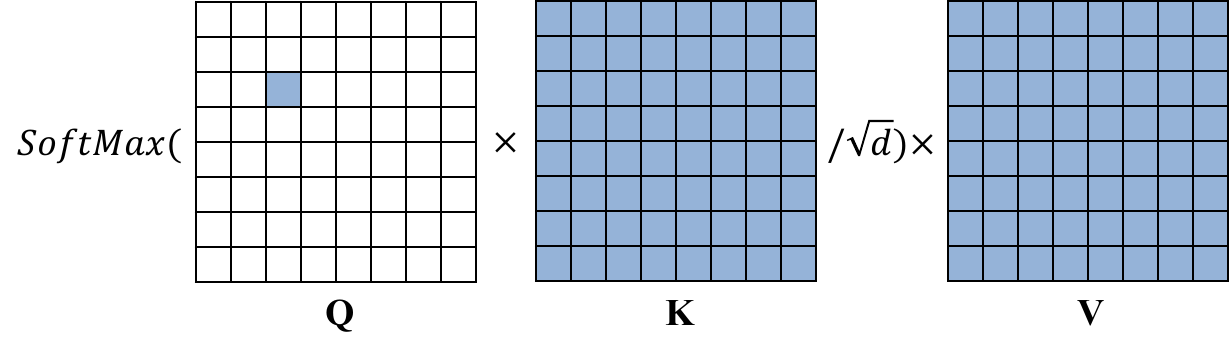}
       \caption{\centering Original window-based self-attention~\cite{liu2021swin}}
	\end{subfigure}\\
	\begin{subfigure}[h]{0.9\linewidth}
		\includegraphics[width=\linewidth]{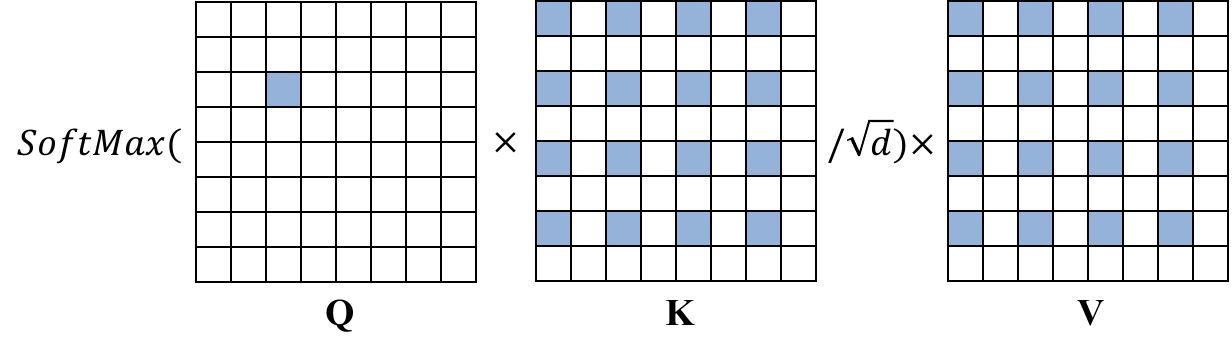}
        \caption{\centering Our masked window-based self-attention (M-WSA)}
	\end{subfigure}
	\caption{Illustration of receptive field. (a) Original window attention performs all-to-all interactions between query and key/value. (b) Our M-WSA applies an elaborate mask to the attention matrix (see Fig.~\ref{fig:overview}(d)), so the blue pixels of query only attend to the locations of key/value at even coordinates, which mimics the receptive field of dilated convolutions.}
    \label{fig:window_attention}
\end{figure}
\subsection{Masked Window-Based Self-Attention (M-WSA)}
Window-based self-attention~\cite{liu2021swin} has been wildly applied in image restoration~\cite{liang2021swinir, chen2023activating}.
In this work, we mimic the behavior of dilated convolutions~\cite{wu2020unpaired, wang2023lg, jang2023puca} to propose a masked window-based self-attention (M-WSA) for SSID, which can be plug-and-played into any layer and exploit current deep features as \textit{query/key/value}.
As shown in Fig.~\ref{fig:overview}(d), we ingeniously design a fixed attention mask adding to the attention matrix to restrict the interactions between \textit{query} and \textit{key/value} tokens.
From Fig.~\ref{fig:window_attention}(a), in original window attention, each \textit{query} token interacts with \textit{key/value} tokens at all spatial locations within the window.
In our M-WSA, the \textit{query} token attends to the spatial locations at even coordinates (see Fig.~\ref{fig:window_attention}(b)).
Therefore, M-WSA exhibits the same functionality as dilated convolutions for building BSN, but with a larger receptive field and stronger local fitting capability.

Here we formally illustrate our attention mask. In window attention, within a local window of size $M\! \times\! M$, the current feature is first projected to \textit{query}, \textit{key} and \textit{value} tokens as $\mathbf{Q}, \mathbf{K}, \mathbf{V}\in\mathbb{R}^{M^2\times d}$, respectively.
Then the original window attention can be formulated as,
\begin{equation}
	\text{Attention}(\mathbf{Q}, \mathbf{K}, \mathbf{V}) = \text{SoftMax}(\mathbf{QK}^T/\sqrt{d})\mathbf{V}.
	\label{eqn:atten1}
\end{equation}
where $d$ is the feature dimension.
In our M-WSA, our attention mask $\mathbf{M}\in\mathbb{R}^{M^2\times M^2}$ is applied to the attention matrix that restricts each \textit{query} only attends to \textit{key/value} at even coordinates, as shown in Fig.~\ref{fig:overview}(d).
Thus, Eq.~(\ref{eqn:atten1}) can be modified as,
\begin{gather}
	\text{Attention}(\mathbf{Q}, \mathbf{K}, \mathbf{V})\! =\! \text{SoftMax}(\mathbf{QK}^T\! /\! \sqrt{d}\! +\! \mathbf{M})\mathbf{V},\\
	\mathbf{M}(i, j)\!=\!\left\{\begin{array}{ll}
		\!0, & \!\text{if}\ x_i\!-\!x_j\!\equiv y_i\!-\!y_j\!\equiv0\ (mod\ 2)\\
		\!-\infty, & \!\text{otherwise}
	\end{array}. \right.
\end{gather}
Specifically, $\mathbf{M}$ is a two-valued matrix that masks out certain locations according to the relative position of \textit{query} (at $i$) and \textit{key/value} (at $j$) tokens.
$(x_i, y_i)$ and $(x_j, y_j)$ are the spatial locations of $i$ and $j$.
When $i$ and $j$ are with even distance on both axes, $\mathbf{M}(i, j)\!=\!0$, the attention value is unchanged.
Otherwise, $\mathbf{M}(i, j)\!=\!-\infty$ and the attention value becomes $0$ after the softmax operation, thereby being masked out.
Inspired by relative position embedding~\cite{liu2021swin}, $\mathbf{M}\in\mathbb{R}^{M^2\times M^2}$ can be calculated from a smaller-sized binary matrix $\hat{\mathbf{M}}\in\mathbb{R}^{(2M-1)\times (2M-1)}$ according to the relative position of $i$ and $j$ to improve the efficiency, \ie,
\begin{equation}
	\mathbf{M}(i, j) = \left\{\begin{array}{ll}
		0, & \text{if}\ \hat{\mathbf{M}}(x_i-x_j, y_i-y_j)=0\\
		-\infty, & \text{if}\ \hat{\mathbf{M}}(x_i-x_j, y_i-y_j)=1
	\end{array}, \right.
\end{equation}
\begin{equation}
	\hat{\mathbf{M}}(x, y) = \left\{\begin{array}{ll}
		0, & \text{if}\ \ x\equiv y\equiv0\ (mod\ 2)\\
		1, & \text{otherwise}
	\end{array} . \right.
\end{equation}
In the implementation perspective, we adopt overlapping cross-attention~\cite{chen2023activating} that calculates \textit{key/value} tokens from a larger field to further expand receptive field.

\begin{figure}
    \centering
    \includegraphics[width=\linewidth]{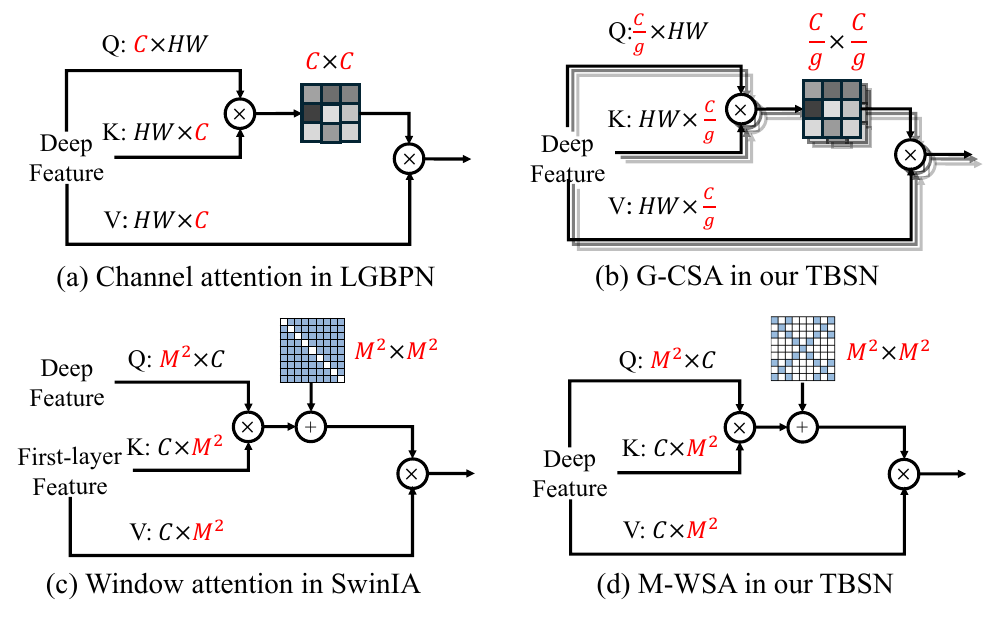}
    \caption{Channel and window attentions in LG-BPN~\cite{wang2023lg} and SwinIA~\cite{papkov2023swinia}. Our G-CSA and M-WSA are distinct from these modules.}
    \label{fig:novelty}
\end{figure}
\textbf{Discussion}. The proposed G-CSA and M-WSA are distinct from the transformer operators in previous BSNs.
As shown in Fig.~\ref{fig:novelty}(a)(b), channel attention in LGBPN~\cite{wang2023lg} is risked to leak blind-spot information when applied in multi-scale architectures, while our G-CSA performs channel attention in separate groups to alleviate this problem.
From Fig.~\ref{fig:novelty}(c)(d), window attention in SwinIA~\cite{papkov2023swinia} masks the main diagonal of the attention matrix to maintain the blind-spot requirement.
Its \textit{key/value} tokens are limited to be from pixel-wise shallow features of the noisy input, thus showing inferior results.
In contrast, our M-WSA applies a dedicated designed mask to mimic the behavior of dilated convolution, which can be flexibly performed on the deep features.

\subsection{Knowledge Distillation for Efficient Inference}
\label{sec:distill}
Self-supervised image denoising methods usually require high computational cost due to complicated network designs~\cite{wu2020unpaired}, increased network size~\cite{jang2021c2n}, and post-processing operation~\cite{lee2022ap}.
The computation burden largely limits their applicability in certain situations, \eg, on mobile devices.
Nonetheless, the performance of SSID methods still falls behind the corresponding supervised ones.
Even lightweight supervised methods may achieve better performance than complex self-supervised ones.
In other words, the lightweight network may be fully sufficient to fit the results from some complex self-supervised methods.
Taking advantage of this, we suggest a knowledge distillation strategy to reduce the inference cost while maintaining the performance.

Specifically, we adopt the efficient U-Net~\cite{ronneberger2015u} architecture as our student network, which is distilled from the self-supervised learned TBSN (namely TBSN2UNet),
\begin{equation}
	\mathcal{L}_{distill} = \left\|sg(\textBF{TBSN}(y)) - \textBF{U-Net}(y)\right\|_1
\end{equation}
where $y$ is the noisy image, $sg(\cdot)$ is the stop gradient operation.
Note that we aim to reduce the computation cost during inference.
It is different from the methods that apply knowledge distillation for better performance~\cite{wu2020unpaired, jang2021c2n, li2023spatially}.

\section{Experiments}
\subsection{Implementation Details}
\noindent\textbf{Datasets}.
We conduct experiments on two wildly used real-world image denoising datasets, \ie, SIDD~\cite{abdelhamed2018sidd} and DND~\cite{plotz2017benchmarking}.
The noisy-clean pairs of SIDD dataset are collected from five smartphone cameras, where each noisy image is captured multiple times and the average image serves as ground truth.
It contains 320 training images, 1280 validation patches and 1280 benchmark patches, respectively.
We train our networks on the noisy images of train split, and test on the benchmark split.
DND is a benchmark dataset collected from DSLR cameras.
The noisy images are captured with a short exposure time while the corresponding clean images are captured with a long exposure time.
It contains 50 pairs for test only.
We train and test our networks on the test images in a fully self-supervised manner.

\noindent\textbf{Training Details}.
For self-supervised training of TBSN, we follow AP-BSN~\cite{lee2022ap} to apply pixel-shuffle downsampling (PD) to break the noise correlation, and adopt asymmetric PD factors during training and inference to trade-off the denoising effect and detail preserving.
We also improve the denoising results with random replacement refinement (R3) strategy.
The batch size and patch size are set to 4 and $128\times 128$, respectively.
We adopt $\ell_1$ loss and AdamW~\cite{loshchilov2018decoupled} optimizer to train the network.
The learning rate is initially set to $3\! \times\! 10^{-4}$, and is decreased by 10 every 40k iterations with total 100k training iterations.
For knowledge distillation, the training settings are the same as self-supervised learning.
All the experiments are conducted on PyTorch framework and Nvidia RTX2080Ti GPUs.

\begin{table}[t]
	\setlength{\tabcolsep}{1mm}
	\small
	\centering
	\begin{tabular}{clcc}
		\toprule
		& \multirow{2}{*}{Method} & SIDD Bench & DND Bench \\
		& & PSNR / SSIM & PSNR / SSIM\\
		\midrule
		Non-learning & BM3D & 25.65 / 0.685 & 34.51 / 0.851\\
		Based & WNNM & 25.78 / 0.809 & 34.67 / 0.865\\
		\midrule
		\multirow{3}{*}{\makecell[c]{Supervised\\(Synthetic pairs)}} & DnCNN & 26.25 / 0.599 & 32.43 / 0.790\\
		& CBDNet & 33.28 / 0.868 & 38.05 / 0.942\\
		& Zhou~\textit{et~al.} & 34.00 / 0.898 & 38.40 / 0.945\\
		\midrule
		\multirow{4}{*}{\makecell[c]{Supervised\\(Real pairs)}} & DnCNN & 37.61 / 0.941 & 38.73 / 0.945\\
		& VDN & 39.26 / 0.955 & 39.38 / 0.952\\
		& Restormer & 40.02 / 0.960 & 40.03 / 0.956\\
		& NAFNet & 40.30 / 0.961 & - / -\\
		\midrule
		\multirow{4}{*}{Unpaired} & GCBD & - & 35.58 / 0.922\\
		& UIDNet & 32.48 / 0.897 & -\\
		& C2N & 35.35 / 0.937 & 37.28 / 0.924\\
		& DBSN &  - & 37.93 / 0.937\\
		\midrule
		\multirow{9}{*}{Self-Supervised} & Noise2Void & 27.68 / 0.668 & -\\
		& Noise2Self & 29.56 / 0.808 & -\\
		& NAC & - & 36.20 / 0.925\\
		& R2R & 34.78 / 0.898 & -\\
		& CVF-SID & 34.71 / 0.917 & 36.50 / 0.924\\
		& AP-BSN & 36.91 / 0.931 & 38.09 / 0.937 \\
		& SASL & 37.41 / 0.934 & 38.18 / 0.938\\
		& LG-BPN & 37.28 / 0.936 & 38.43 / \underline{0.942} \\
		& PUCA & 37.54 / 0.936 & 38.83 / \underline{0.942} \\
            & AT-BSN & \underline{37.78} / \textBF{0.944} & 38.68 / \underline{0.942} \\
        \midrule
	\multirow{2}{*}{\makecell[c]{Self-Supervised\\(Ours)}}	& TBSN & \underline{37.78} / \underline{0.940} & \textBF{39.08} / \textBF{0.945}\\
		& TBSN2UNet & \textBF{37.79} / \underline{0.940} & \underline{39.01} / \textBF{0.945} \\
		\bottomrule
	\end{tabular}
    \caption{Quantitative comparison on SIDD and DND benchmark datasets. In self-supervised category, the first and second place results are highlighted in \textbf{bold} and \underline{underlined}.}
	\label{tab:sidd_dnd}
\end{table}
\begin{figure}[t]
	\newcommand{\siddsubfigurelen}{0.24}
	\centering
	\begin{subfigure}[b]{\siddsubfigurelen\linewidth}
		\centering
		\includegraphics[width=\linewidth]{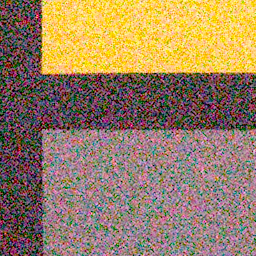}
		\caption*{Noisy}
	\end{subfigure}
	\hfill
	\begin{subfigure}[b]{\siddsubfigurelen\linewidth}
		\centering
		\includegraphics[width=\linewidth]{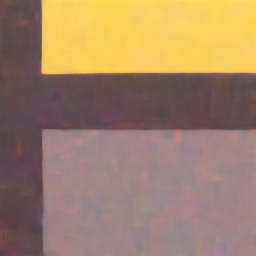}
		\caption*{AP-BSN}
	\end{subfigure}
	\hfill
	\begin{subfigure}[b]{\siddsubfigurelen\linewidth}
		\centering
		\includegraphics[width=\linewidth]{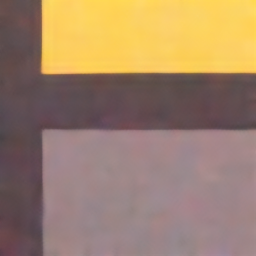}
		\caption*{LG-BPN}
	\end{subfigure}
	\hfill
	\begin{subfigure}[b]{\siddsubfigurelen\linewidth}
		\centering
		\includegraphics[width=\linewidth]{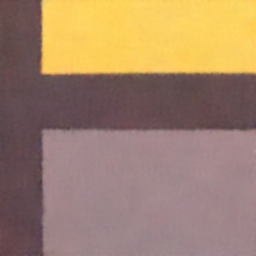}
		\caption*{SASL}
	\end{subfigure}\\
	\hfill
	\begin{subfigure}[b]{\siddsubfigurelen\linewidth}
		\centering
		\includegraphics[width=\linewidth]{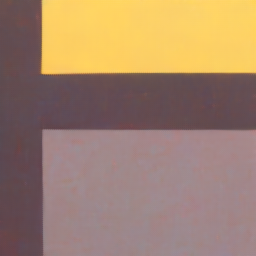}
		\caption*{PUCA}
	\end{subfigure}
	\hfill
	\begin{subfigure}[b]{\siddsubfigurelen\linewidth}
		\centering
		\includegraphics[width=\linewidth]{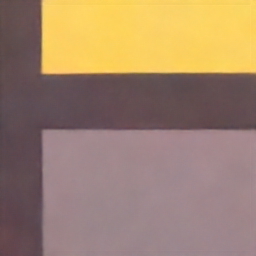}
		\caption*{AT-BSN}
	\end{subfigure}
	\hfill
	\begin{subfigure}[b]{\siddsubfigurelen\linewidth}
		\centering
		\includegraphics[width=\linewidth]{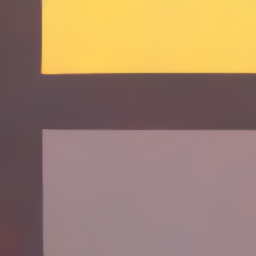}
		\caption*{TBSN(Ours)}
	\end{subfigure}
	\hfill
	\begin{subfigure}[b]{\siddsubfigurelen\linewidth}
		\centering
		\includegraphics[width=\linewidth]{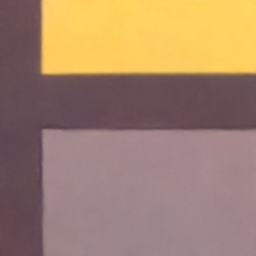}
		\caption*{TBSN2UNet}
	\end{subfigure}\\
	\caption{ Qualitative comparison on SIDD dataset, please zoom in for better observation.}
	\label{fig:sidd}
\end{figure}
\begin{figure}[t]
	\newcommand{\dndsubfigurelen}{0.24}
	\centering
	\begin{subfigure}[b]{\dndsubfigurelen\linewidth}
		\centering
		\includegraphics[width=\linewidth]{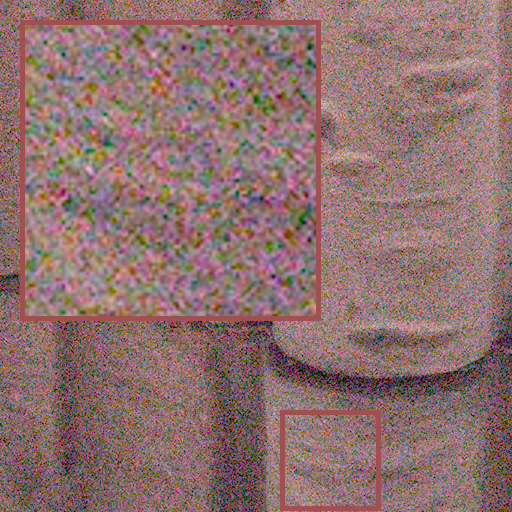}
            \caption*{Noisy}
	\end{subfigure}
	\hfill
 	\begin{subfigure}[b]{\dndsubfigurelen\linewidth}
		\centering
		\includegraphics[width=\linewidth]{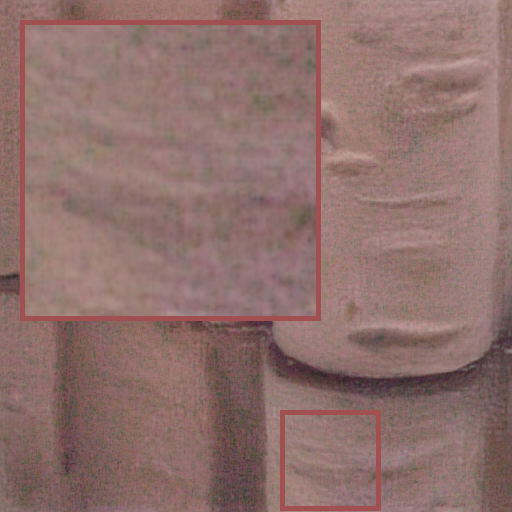}
            \caption*{C2N}
	\end{subfigure}
	\hfill
	\begin{subfigure}[b]{\dndsubfigurelen\linewidth}
		\centering
		\includegraphics[width=\linewidth]{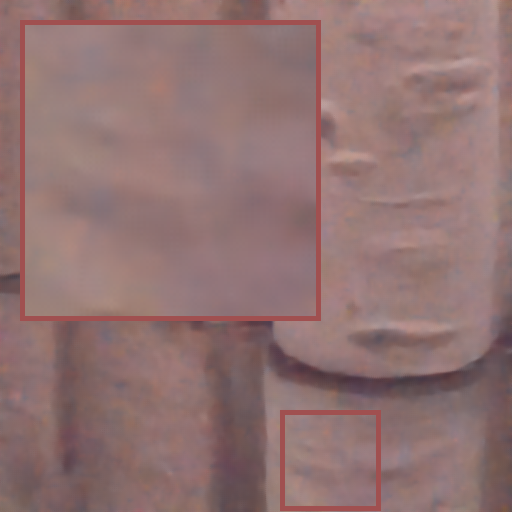}
            \caption*{AP-BSN}
	\end{subfigure}
		\hfill
	\begin{subfigure}[b]{\dndsubfigurelen\linewidth}
		\centering
		\includegraphics[width=\linewidth]{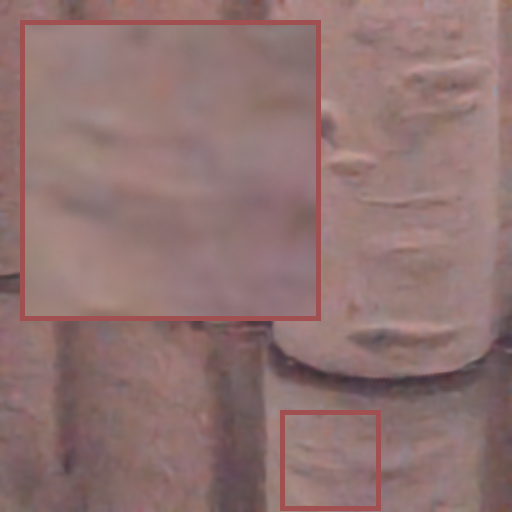}
		\caption*{LG-BPN}
	\end{subfigure}\\
	\hfill
	\begin{subfigure}[b]{\dndsubfigurelen\linewidth}
		\centering
		\includegraphics[width=\linewidth]{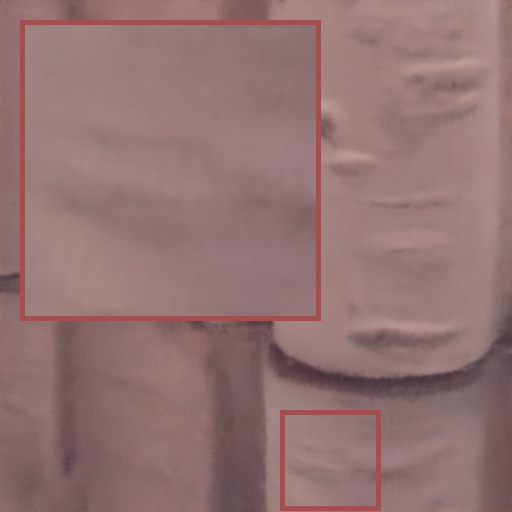}
            \caption*{SASL}
	\end{subfigure}
	\hfill
	\begin{subfigure}[b]{\dndsubfigurelen\linewidth}
		\centering
		\includegraphics[width=\linewidth]{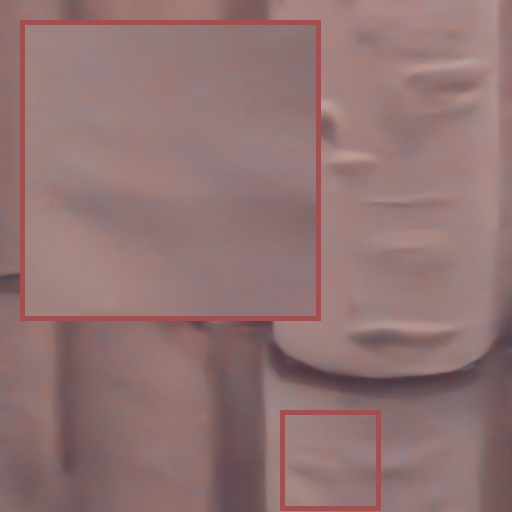}
            \caption*{PUCA}
	\end{subfigure}
	\hfill
	\begin{subfigure}[b]{\dndsubfigurelen\linewidth}
		\centering
		\includegraphics[width=\linewidth]{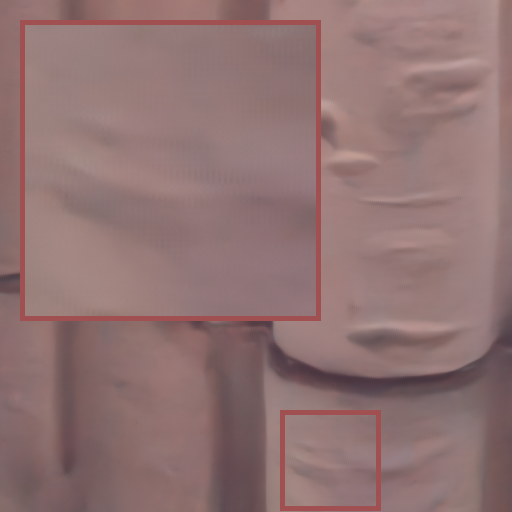}
            \caption*{TBSN (Ours)}
	\end{subfigure}
	\hfill
	\begin{subfigure}[b]{\dndsubfigurelen\linewidth}
		\centering
		\includegraphics[width=\linewidth]{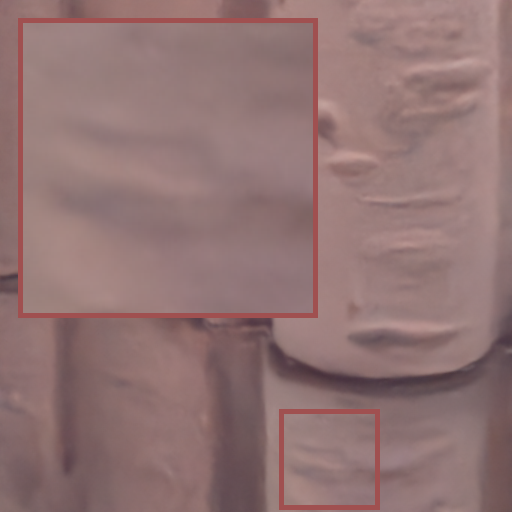}
		\caption*{TBSN2UNet}
	\end{subfigure}\\
	\caption{ Qualitative comparison on DND dataset.}
    \label{fig:dnd}
\end{figure}
\subsection{Comparison with State-of-the-Art Methods}
\noindent\textbf{Quantitative comparison}.
Tab.~\ref{tab:sidd_dnd} shows the quantitative results of proposed TBSN and state-of-the-art self-supervised methods: Noise2Void~\cite{krull2019noise2void}, Noise2Self~\cite{batson2019noise2self}, NAC~\cite{xu2020noisy}, R2R~\cite{pang2021recorrupted}, CVF-SID~\cite{neshatavar2022cvf}, AP-BSN~\cite{lee2022ap}, SASL~\cite{li2023spatially}, LG-BPN~\cite{wang2023lg}, PUCA~\cite{jang2023puca} and AT-BSN~\cite{chen2024exploring}.
Among these, blind-spot techniques designed for spatial independent noise (Noise2Void, Noise2Self, and R2R) exhibit little denoising effect on real-world noisy images. 
Although pixel-shuffle downsampling (PD) breaks the noise correlation and successfully removes the noise~\cite{lee2022ap}, the performance is still limited by its plain convolutional BSN architecture.
Some recent works tackle this problem by searching for advanced BSN architectures.
For instance, LG-BPN incorporates transformer block~\cite{zamir2022restormer} into BSN for global information and brings 0.37dB improvement over AP-BSN baseline.
PUCA designs a multi-scale BSN with channel attention and achieves 0.63dB improvement.
Nonetheless, benefitting from channel and window attention mechanisms, our TBSN boosts the improvement to 0.87dB on SIDD benchmark dataset.
AT-BSN introduces multi-teacher distillation strategy, in comparison, our TBSN2UNet achieves 0.01dB and 0.33dB improvement against AT-BSN in SIDD and DND Benchmark datasets respectively, showing the superiority of our method.

\noindent\textbf{Qualitative comparison}.
The qualitative results of self-supervised image denoising methods are shown in Fig.~\ref{fig:sidd} and Fig.~\ref{fig:dnd}.
The denoising of the color chart in Fig.~\ref{fig:sidd} depends on the global information, where former methods fail to wipe the noise completely.
Benefiting from the channel-wise self-attention, our TBSN removes the spatially correlated noise smoothly.
The cup and the wall in Fig.~\ref{fig:sidd} and Fig.~\ref{fig:dnd} show that TBSN could maintain the details due to its local fitting capability of window attention.

\subsection{Comparison of Model Complexity}
The channel-wise~\cite{zamir2022restormer} and window-based~\cite{liu2021swin} self-attentions in TBSN are efficient transformer modules designed for image restoration task.
In addition, TBSN adopts hierarchical multi-scale architecture to further improve its efficiency.
As shown in Tab.~\ref{tab:complexity}, TBSN maintains similar computation complexity as the convolutional counterparts PUCA, and is more efficient than LG-BPN.
In addition, SASL shows attractive \#Param and \#FLOPs results due to its U-Net architecture.
For a fair comparison, the U-Net distilled from our TBSN exhibits the same computation complexity as SASL but with higher performance, which demonstrates the superiority of our knowledge distillation strategy.

\section{Ablation Study}
\subsection{Visualization of the Receptive Field}
The expansion of receptive field is a major factor towards the success of transformers~\cite{zamir2022restormer}.
In this subsection, we plot the input pixels for recovering the center pixel of output to access the effective receptive field of TBSN.
Specifically, we pass an input image through the network, select the center pixel of the output image, and calculate its gradient with respect to the input image.
The pseudo code in PyTorch format is as follows,
\begin{lstlisting}[language=Python, numbers=none]
import torch as t

image = t.rand((1,3,H,W),requires_grad=True)
output = model(image)
center_pixel = t.mean(output[...,H//2,W//2])
center_pixel.backward()
gradient = t.sum(t.abs(x.grad),dim=1,keepdim=True)
\end{lstlisting}
Such gradient indicates how much the output pixel changes with a disturbance on each input pixel.
We sum the absolute value of the gradient along the channel axis for visualization.
From Fig~\ref{fig:receptive_field}, TBSN shows a significantly wider receptive field than former BSNs~\cite{lee2022ap, wang2023lg, jang2023puca}, which is a possible explanation for the appealing performance of TBSN.

\begin{table}[t]
	\setlength{\tabcolsep}{1mm}
	\small
	\centering
	\begin{tabular}{cccccc}
		\toprule
		Method & SASL & LG-BPN & PUCA & TBSN & TBSN2UNet \\
		\midrule
		PSNR (dB) & 37.41 & 37.28 & 37.54 & 37.78 & 37.79 \\
		\#Param (M) & 1.08 & 6.61 & 12.78 & 12.97 & 1.08 \\
		\#FLOPs (G) & 35.0 & 6699.6 & 2644.2 & 5463.9 & 35.0 \\ 
		Time (ms) & 4.8 & 5229.4 & 444.7 & 1032.1 & 4.8 \\
		\bottomrule
	\end{tabular}
    \caption{Model complexity comparison on SIDD benchmark dataset. \#FLOPs and time are measured on $256\!\times\! 256$ size.}
	\label{tab:complexity}
\end{table}
\subsection{Analysis on DTAB}
\label{sec:ablation_dtab}
Tab.~\ref{tab:dtab} analyzes the effectiveness of the components in our dilated transformer attention block (DTAB).
We begin with a base model (1) that degenerates the window and channel attentions to dilated convolutions~\cite{wu2020unpaired}.
In contrast, our dilated M-WSA (2) enhances the base model with local fitting capability and provides 0.27dB improvement.
Our dilated G-CSA (3) exhibits global interaction and shows 0.65dB improvement.
TBSN achieves a total improvement of 0.81 with combined channel and window attentions, which demonstrates the complementarity of the local and global operations.
In addition, we assess the effects of other window and channel attention implementations in Fig.~\ref{fig:novelty}.
Replacing M-WSA with SwinIA (4) leads to 0.59dB performance drop while replacing G-CSA with LG-BPN (6) leads to 0.05dB performance drop.
The other attention mechanisms, Swin Transformer~\cite{liu2021swin} (5), SE~\cite{hu2018squeeze} (7) and SCA~\cite{chen2022simple} (8) also shows inferior performance.
To summarize, the ablation study results suggest that DTAB is the optimal network choice.

\begin{table}[t]
	\setlength{\tabcolsep}{2.5mm}
	\small
	\centering
	\begin{tabular}{cccc}
		\toprule
		Model & Window Attention & Channel Attention & PSNR\\
		\midrule
		(1) & - & - & 36.90 \\
		(2) & Dilated M-WSA & - & 37.17 \\
		(3) & - & Dilated G-CSA & 37.55 \\
		\midrule
		(4) & SwinIA & Dilated G-CSA & 37.12 \\
            (5) & Swin Trans. & Dilated G-CSA & 37.56\\
            \midrule
		(6) & Dilated M-WSA & LG-BPN & 37.66 \\
            (7) & Dilated M-WSA & SE & 37.43 \\
            (8) & Dilated M-WSA & SCA & 37.63 \\
		\midrule
		TBSN & Dilated M-WSA & Dilated G-CSA & 37.71 \\
		\bottomrule
	\end{tabular}
    \caption{Ablation study of dilated transformer attention block (DTAB) on SIDD validation dataset. '-' refers to plain dilated convolutions. (1$\sim$3) indicates our dilated M-WSA and G-CSA are essential components of DTAB. (4$\sim$8) shows they are not replaceable with analogous operators.}
	\label{tab:dtab}
\end{table}
\subsection{Analysis on Channel Attention}
As illustrated in the method, channel-wise self-attention (CSA)~\cite{wang2023lg} may leak blind-spot information in multi-scale architectures.
Tab.~\ref{tab:ca} analyzes the effect of CSA with downsampling scales from 1 to 5.
In correspondence, the channel number at the deepest layers grows from 48 to 768, and the spatial resolution reduces from $128^2$ to $8^2$, respectively.
From the middle lines of Tab.~\ref{tab:ca}, the plain dilated CSA provides positive effects at the scales less equal to 3, but leads to obvious performance drop at the 4- and 5-scales.
This is owing to the channel dimension being larger than the spatial resolution at 4- and 5-scales so it leaks the blind-spot information.
Instead, our grouped dilated G-CSA divides the channels into several groups and performs channel attention separately.
As the channel dimension within each group is controlled smaller than the spatial resolution, dilated G-CSA provides constant improvement at all scales.

\subsection{Analysis on Knowledge Distillation}
As shown in Tab.~\ref{tab:distill}, we conduct experiments with U-Net architecture to assess the effectiveness of knowledge distillation.
Despite fewer parameters and FLOPs, U-Net trained in a supervised manner achieves 1.21dB higher performance than TBSN on SIDD validation dataset. It demonstrates lightweight U-Net has enough learning capacity to receive the denoising performance of TBSN.
Consequently, the distilled student U-Net achieves comparable results as the teacher TBSN.
The results show that knowledge distillation is a feasible way to reduce the model size and computation cost during inference in SSID.

\begin{table}[t]
	\setlength{\tabcolsep}{2mm}
	\small
	\centering
	\begin{tabular}{cccccc}
		\toprule
		\#Scale & 1 & 2 & 3 & 4 & 5\\
		\midrule
		Max. \#Channel & 48 & 96 & 192 & 384 & 768 \\
		Min. \#Spatial & 128$^2$ & 64$^2$ & 32$^2$ & 16$^2$ & 8$^2$ \\
		w/o Dilated CSA & 37.15 & 37.29 & 37.55 & 37.07 & 37.02 \\
		w/ Dilated CSA & 37.33 & 37.51 & 37.66 & 36.84 & 11.35 \\
		\midrule
		\#Group & 1 & 2 & 4 & 8 & 16 \\
		\#Channel / \#Group & 48 & 48 & 48 & 48 & 48 \\
		w/ Dilated G-CSA & 37.34 & 37.52 & 37.71 & 37.38 & 37.31 \\
		\bottomrule
	\end{tabular}
    \caption{Ablation study of channel-wise self-attention (CSA) on SIDD validation dataset. Applying CSA to the 4- or 5-scale architectures causes performance drop as it leaks the blind-spot information. Our grouped CSA (G-CSA) eliminates the negative effects by splitting the channels into groups and performing CSA separately.}
	\label{tab:ca}
\end{table}
\begin{table}[t]
	\setlength{\tabcolsep}{0.9mm}
	\small
	\centering    
	\begin{tabular}{llcccc}
		\toprule
		\multirow{2}{*}{Training}& \multirow{2}{*}{Model} & PSNR & \#Param & \#FLOPs & Time \\
		& & (dB) & (M) & (G) & (ms) \\
		\midrule
		Supervised & Network: U-Net & 38.92 & 1.08 & 35.0 & 4.8 \\
		\midrule
		Knowledge & Teacher: TBSN & 37.71 & 12.97 & 607.1 & 1032.1 \\
		Distillation & Student: U-Net & 37.70 & 1.08 & 35.0 & 4.8 \\
		\bottomrule
	\end{tabular}
    \caption{Ablation study of knowledge distillation on SIDD validation dataset.}
	\label{tab:distill}
\end{table}
\section{Conclusion}
In this paper, we propose a transformer-based blind-spot network, namely TBSN, for self-supervised image denoising.
Key designs are introduced to adapt the spatial and channel self-attention operators for constructing BSN.
For spatial attention, an elaborate mask is applied to the window attention matrix, thus restricting its receptive field to mimic the dilated convolutions.
For the spatial information leakage problem of channel attention, we propose to perform channel attention in separate groups to eliminate its harmful effects.
Moreover, a knowledge distillation strategy is introduced to reduce the computation cost during inference.
Extensive experiments on real-world denoising datasets demonstrate that TBSN largely expands the effective receptive field and achieves state-of-the-art performance.

\section{Acknowledgements}
This work was supported in part by the National Natural Science Foundation of China (NSFC) under Grants 62371164 and U22B2035.

\bibliography{aaai25}

\end{document}